\documentclass[preprint,12pt]{elsarticle}

\usepackage{amssymb}

\usepackage{amsmath,amssymb,amsfonts}

\usepackage{algorithmic}
\usepackage{graphicx}
\usepackage{textcomp}
\usepackage{xcolor}
\usepackage{booktabs}
\usepackage{multirow}
\usepackage{tabularray}
\usepackage[framemethod=tikz]{mdframed}
\usepackage{spverbatim}
\usepackage{hyperref}
\usepackage{threeparttable}
\UseTblrLibrary{booktabs}
\usepackage[T1]{fontenc}

\definecolor{light-gray}{gray}{0.95}

\usepackage{subcaption}
\usepackage{rotating} %
\usepackage{listings}

\journal{Information Systems}

\newif\ifrevision
\revisiontrue  %

\newcommand{\revise}[1]{%
  \ifrevision
    {\color{black}#1}%
  \else
    #1%
  \fi
}

\begin{document}

\begin{frontmatter}

\title{SMUTF: Schema Matching Using Generative Tags and Hybrid Features}

\author[ncu]{Yu Zhang$^*$}
\author[nwu]{Di Mei$^*$}
\author[nwu]{Haozheng Luo$^*$}
\author[nwu]{Chenwei Xu}
\author[ncu,as]{Richard Tzong-Han Tsai$^*{}^*$}

\affiliation[ncu]{
  organization={Department of Computer Science and Information Engineering, National Central University},
  addressline={No. 300, Zhongda Rd, Zhongli District},
  postcode={320},
city={Taoyuan},
state={Taiwan}
}

\affiliation[nwu]{%
  organization={Department of Computer Science, Northwestern University},
  addressline={633 Clark St},
  postcode={60208},
  city={Evanston},
  state={IL,USA}
}

\affiliation[as]{%
  organization={Center for Geographic Information Science, Research Center for Humanities and Social Sciences, Academia Sinica},
  addressline={128 Academia Road},
  postcode={115201},
  city={Taipei},
  state={Taiwan}
}

\begin{abstract}
We introduce \textbf{SMUTF} (\textbf{S}chema \textbf{M}atching \textbf{U}sing Generative \textbf{T}ags and Hybrid \textbf{F}eatures), a unique approach for large-scale tabular data schema matching (SM), which assumes that supervised learning does not affect performance in open-domain tasks, thereby enabling effective cross-domain matching. This system uniquely combines rule-based feature engineering, pre-trained language models, and generative large language models. In an innovative adaptation inspired by the Humanitarian Exchange Language, we deploy "generative tags" for each data column, enhancing the effectiveness of SM. SMUTF exhibits extensive versatility, working seamlessly with any pre-existing pre-trained embeddings, classification methods, and generative models. 

Recognizing the lack of extensive, publicly available datasets for SM, we have created and open-sourced the HDXSM dataset from the public humanitarian data. We believe this to be the most exhaustive SM dataset currently available. In evaluations across various public datasets and the novel HDXSM dataset, SMUTF demonstrated exceptional performance, surpassing existing state-of-the-art models in terms of  accuracy and efficiency, and improving the F1 score by 11.84\% and the AUC of ROC by 5.08\%. \textcolor{black}{Code is available at \href{https://github.com/fireindark707/Python-Schema-Matching}{GitHub}}.

\end{abstract}

\begin{graphicalabstract}
\end{graphicalabstract}

\begin{highlights}
\item The development of SMUTF, a novel Schema Matching system surpassing current state-of-the-art methods.
\item The demonstration of generative language models' beneficial use in improving Schema Matching.
\item The recognition of supervised learning's ability to strengthen Schema Matching across a wide variety of datasets.
\item The creation of HDXSM, a new, comprehensive Schema Matching dataset.
\end{highlights}


\end{frontmatter}
\def\thefootnote{*}\footnotetext{These authors contributed equally to this work}\def\thefootnote{\arabic{footnote}}
\def\thefootnote{**}\footnotetext{Corresponding Author}\def\thefootnote{\arabic{footnote}}

\section{Introduction}

Open Data, characterized by its accessibility, exploitability, and shareability, is seen as an effective means for governments and Non-Governmental Organizations (NGOs) to enhance public services and accountability. Despite the promising potential, challenges abound due to the distributed and heterogeneous nature of data environments and the diversity in data handling methodologies.
Dataset discovery, a key component of the data integration process \cite{doan2012principles} in today's intricate data environment, hinges on schema matching (SM) \cite{Koutras2020ValentineEM} — a technique for unraveling the relationships between columns (elements) under different tabular schemas. The effectiveness of SM plays a critical role in determining the quality of dataset discovery, which in turn affects the performance of further data analysis.

\begin{figure}[h]
    \centering
    \includegraphics[width=\textwidth]{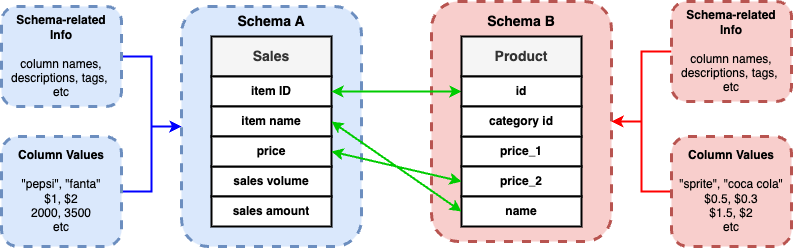}
    \caption{Schema matching aims to discover relationships between columns.}
    \label{fig:schema_match_intro}
\end{figure}

This paper introduces a novel approach to SM, termed as \textbf{SMUTF} (\underline{\textbf{S}}chema \underline{\textbf{M}}atching \textbf{U}sing Generative \textbf{T}ags and Hybrid \textbf{F}eatures). SMUTF is designed to perform SM on any type of tabular data with improved accuracy. \textcolor{black}{Most traditional SM approaches~\cite{Koutras2020ValentineEM,madhavan2001generic,melnik2002similarity,Do2002COMAA,Zhang2011AutomaticDO} are sensitive to ad hoc SM scenarios, and they are inclined to depend on either the schema-related information (column names, descriptions, etc) or the column values (column instances), as shown in Figure~\ref{fig:schema_match_intro}. In contrast to them, SMUTF utilizes supervised learning, thereby enhancing the robustness of SM under various conditions, a claim substantiated by our experimental results. \revise{Though many contemporary works, especially those employing deep learning algorithms \cite{traeger2025collective, miazga2025automated, ma2025knowledge, seedat2025matchmaker, zhang2023schema, shraga2022powarematch}, have adopted pre-trained transformer-based models supervised fine-tuned on matched data pairs when addressing the problem, they are limited to many issues from real-world data processing situations like varying column formats and matches between less-relevant tables or confusing columns.} For instance, it's really hard to differentiate "price\_1" (price at purchase) from "price\_2" (sales price) in Figure~\ref{fig:schema_match_intro} purely based on their names and values, and this requires a more complex matching mechanism exploring as well as involving different types of table information to overcome the challenge. The well-designed integration of text embeddings, rule-based features and generated descriptive tags proposed by our approach can eliminate these obstacles.}

SMUTF combines the strength of Pretrained Language Models (PLM) \cite{devlin2018bert, liu2019roberta, he2020deberta, ouyang2022training} to generate descriptive tags of data under each column and semantic embeddings on textual information. Further, similarity measures  between schemas are computed considering various factors such as dataset values, column names, generated tags, etc. These combined features are used within a gradient-boosting model, XGBoost \cite{chen2016xgboost}, to predict  column matches across datasets.

Also, \textcolor{black}{one prevalent issue in the SM field is the scarcity of non-synthetic open-sourced evaluation datasets \cite{10.1145/3588938}. Some open-sourced SM datasets are not intact, with only either the schema information or the attribute values; others, though complete in terms of their contents, are not complex enough for nowadays' SM evaluations since they only contain few tables, attributes and values.} Thus, we have taken the initiative to create the \textbf{HDXSM} dataset. The data used in this dataset is derived from the Humanitarian Data Exchange (HDX) and is annotated with existing HXL tags \cite{Keler2015TheHE}. The annotation process involves extensive manual checks and verification to ensure dataset's validity \textcolor{black}{\cite{shraga2022humanalcalibratinghumanmatching}}. As far as we are aware, the \textbf{HDXSM} dataset represents the most comprehensive resource currently  available in this field. 

Experiments performed using SMUTF demonstrate significant improvements over existing techniques. Our results indicate enhancements of 11.84\% and 5.08\% in the model's F1 and AUC scores, respectively.

The primary contributions of our paper are summarized:

\begin{enumerate}
\item \textcolor{black}{The development of SMUTF, a novel SM system outperforming current state-of-the-art SM methods.}
\item \textcolor{black}{An exhaustive evaluation showing the ability of supervised learning with hybrid features and generated tags to strengthen SM across a wide variety of domains.}
\item \textcolor{black}{The demonstration of generative language models' beneficial use to improve the performance of SM.}
\item \textcolor{black}{The creation of HDXSM dataset, a new and comprehensive SM dataset.}
\end{enumerate}

The paper subsequently presents a review of related work, provides a detailed introduction of our approach, showcases the experimental results, and concludes with potential future directions for this line of research.

\section{Related work}
Our research is interlinked with three main fields: Schema Matching (SM), Text Embedding with Pretrained Language Models (PLM), and  Metadata Generation with Large Language Models (LLM).

\begin{table*}[h]
\centering
\caption{Schema Matching Component \cite{Koutras2020ValentineEM}, in which CN represents Column Name }
\footnotesize
\resizebox{\textwidth}{!}{%
\begin{tblr}{ X[0.5,l]  X[0.4,l] X[1.7,l]  X[0.8,l] } 
\toprule
\textbf{Matcher Type} & \textbf{Focuses on} & \textbf{Two Columns are Related When...} & \textbf{Corresponding Component in SMUTF} \\ 
\midrule
Attribute Overlap & CN & A syntactic overlap above a given threshold & Column Name Feature Extraction\\ 
\hline
Value Overlap & Values & Corresponding value sets significantly overlap & - \\ 
\hline
Semantic Overlap & CN, Values & A significant overlap between the derived labels using an external knowledge base & HXL-style Tag Generation \\ 
\hline
Data Type & Values & Share same data type (integer, string, etc.) & Value Feature Extraction \\ 
\hline
Distribution & Values & Share similar distributions & Value Feature Extraction \\ 
\hline
Embeddings &CN, Values & Similarity of the embeddings is high & Deep Embedding Cosine Similarity \\
\bottomrule
\end{tblr}
}
\label{table:component}
\end{table*}
\vspace{-0.1in}
\subsection{Schema Matching}

SM and data integration critically depend on measuring similarity and understanding data diffusion across systems, as explored in Fernandez's study of "data seeping" \cite{fernandez2018seeping} and its role in SM efficiency. Melnik's work \cite{melnik2002similarity} on similarity measures and Madhavan's universal SM model \cite{madhavan2001generic} have advanced the field, but evolving schemas and database complexities present ongoing challenges. Traditional machine learning and hashing methods often struggle with complex matchings, as seen in Courps' \cite{madhavan2005corpus} focus on simple attribute alignment. Recent efforts have incorporated neural networks  \cite{zhang2023schema, shraga2022powarematch,shraga2020adnev,li2020ditto,zhang2021smat}, using pre-trained models and LSTM architectures for SM, yet these approaches can falter with variable-length sequences and long-distance relationships within data. For example, DITTO \cite{li2020ditto} leverages \revise{Pretrained Language Models (PLMs)} for entity matching by employing a fine-tuned BERT model to process pairs of records and predict their match likelihood. DITTO's use of domain-specific augmentations and task-specific pretraining highlights the potential of PLMs to adapt to structured and semi-structured data. While SMUTF shares similarities with DITTO, such as leveraging PLMs for semantic understanding, our work extends beyond pairwise matching by incorporating HXL-style tags and hybrid features for schema alignment across datasets.
\vspace{-0.1in}

\revise{

\subsection{Text Embedding with PLM}

In recent years, Transformer-based PLMs have achieved significant success in various Natural Language Processing (NLP) tasks \cite{luo2020openended,liu2022sciannotate,qin2021ibert}. These models utilize self-supervised learning methods during the pre-training phase, including the Cloze task, where the model is trained to predict masked parts of a sentence. However, our study primarily focuses on text embedding using subword tokenization and sentence embedding. We opt for sentence embeddings as they encapsulate sentence-level semantics and reduce the dimensionality, offering efficient training and quicker inference time compared to word embeddings. Rather than utilizing traditional bag-of-words (BoW) models \cite{pagliardini2017unsupervised} or the skip-thought model \cite{kiros2015skipthought}, our research employs transformer-based models for sentence embedding. These models \cite{Reimers2019SentenceBERTSE, feng2022languageagnostic, cer2018universal} exploit positional encoding in the attention mechanism \cite{vaswani2017attention}, which aids in understanding the interrelationships between words in a sentence. This feature is crucial in comprehending sentence-level semantics. Furthermore, the self-attention mechanism within these models assigns weight to each word in a sentence based on its relationship with other words, enabling the transformer to capture the sentence's meaning more accurately.
\vspace{-0.1in}

\subsection{LLM-Based Approaches for Tabular Data}
Large Language Models (LLMs) represent an exciting frontier for tabular data understanding \cite{pan2024chain,sui2024table,xu2024bishop,li2023table}. Table-GPT \cite{li2023table}, for instance, investigates how GPT-style models can process tabular data directly for tasks like table completion and semantic type annotation. CoA \cite{pan2024chain} introduces a chain-of-thought approach for processing tabular data and generating answers to questions derived from the provided data.  While most of LLM-based tabular methods \cite{li2023table, pan2024conv} primarily focus on table reasoning and completion, SMUTF integrates LLMs to generate domain-specific annotations, demonstrating their utility in schema alignment.
}

\subsection{Metadata Generation with LLM}

Large Language Models (LLM) have become pivotal in NLP and machine learning research due to their multifaceted applications. Most of works  \cite{NEURIPS2020_1457c0d6,raffel2020exploring,touvron2023llama, zhang2022opt} are designed to generate accurate descriptions of given inputs and have been used in a range of tasks, including question-answering, summarization, content creation, and translation. Despite these advancements, the potential of LLMs in summarizing and describing tabular data remains relatively unexplored. 
Our approach aims to fill this gap by utilizing LLMs to understand provided data such as column names and values. Subsequently, we leverage the auto-regressive property of Transformers to generate descriptive, Humanitarian Exchange Language style (HXL-style) tags for the columns. These tags offer a high-level synopsis or classification of the assigned data, thereby allowing for a succinct understanding of the data's content. Our innovative application of LLMs demonstrates their potential in the realm of tabular data summarization and description.

\subsection{Semantic Type Detection and Table Understanding}
Semantic type detection \cite{hulsebos2023adatyper,hulsebos2019sherlock,10.14778/3407790.3407793} and table understanding \cite{deng2022turl,arik2021tabnet} are foundational tasks for schema matching and column annotation. These tasks aim to infer the semantic meaning of columns in tabular data, providing critical insights that facilitate schema alignment and integration.

Sherlock \cite{hulsebos2019sherlock} combines handcrafted statistical features, pre-trained embeddings, and neural networks to classify columns into a fixed set of semantic types. Sherlock's capability to identify types such as "Location," "Date," or "Currency" is particularly useful for schema matching, as it provides structured labels that assist in aligning columns across disparate schemas. Unlike Sherlock, SMUTF’s tagging approach is more adaptable and capable of generating tags for open-domain and humanitarian-specific data.
Other works in semantic table annotation, such as TabNet \cite{arik2021tabnet} and TURL \cite{deng2022turl}, emphasize the integration of schema-level information with data-level signals. These approaches use transformer-based architectures to extract context-sensitive representations of column data, improving the precision of column type inference. 

SMUTF builds on these advancements by integrating semantic type detection into its hybrid approach. By generating HXL-style tags that combine semantic understanding with domain-specific annotations, SMUTF enhances column annotation and improves schema matching performance across diverse datasets.

\begin{figure}[h]
    \centering
    \includegraphics[width=\textwidth]{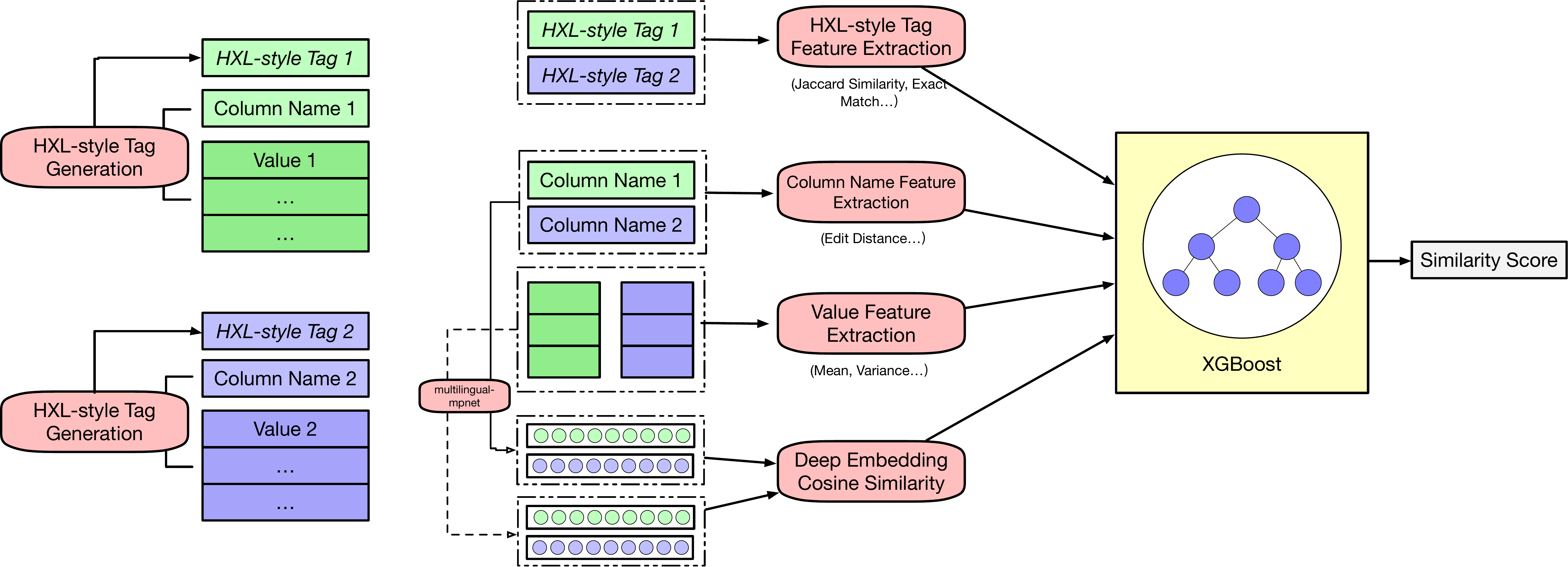}
    \caption{The basic design of SMUTF comprises two primary elements: the generation of HXL-style tags and the calculation of similarity. Four additional computations are employed for measuring similarity. The outcome of these computations, the similarity score, is then used to to predict if two columns are a match.}
    \label{fig:schema}
\end{figure}

\section{SMUTF Methodology}
The SM strategy proposed in this paper, termed \textbf{SMUTF} (\underline{\textbf{S}}chema \underline{\textbf{M}}atching \underline{\textbf{U}}sing Generative \underline{\textbf{T}}ags and Hybrid \underline{\textbf{F}}eatures), consists of four components: \textit{HXL-style tags generation}, \textit{rule-based feature extraction}, \textit{deep embedding similarity}, and \textit{similarity score prediction using XGBoost}.
\vspace{-0.15in}
\subsection{Problem Definitions}

Our primary objective is to devise an SM methodology that can independently establish the relationship between two distinct schemas, aligning their respective columns with the assistance of machine learning models. The proposed method \textcolor{black}{does not require any external knowledge base to assist the SM process and it} consists of two fundamental tasks: \textit{generating HXL-style tags} and \textit{calculating similarity scores}.

We perceive our schema-matching task as a problem of similarity matching (as illustrated in Figure~\ref{fig:schema}) involving two  \textcolor{black}{schemas} $S = \left \langle C,V \right \rangle$. In this context, a schema is made up of column names $C = \{c_1, c_2, \dots, c_n\}$ and corresponding values $V = \{v_{i,1}, v_{i,2}, \dots, v_{i,m}\}$ for the i-th column where $i \leq n$. Essentially, the names and values of each column are viewed as a sequence labeling problem, for which the Large Language Model (LLM) generates HXL-style tags.  \textcolor{black}{These tags are} merged with other column features and  \textcolor{black}{then merged} features are fed into a gradient-boosting method, XGBoost, to perform classification. This results in the prediction of the similarity score between two columns. For every column set, either from the source $S_{src}$ or the target schema $S_{tar}$, the model's outcome is a similarity score matrix  \textcolor{black}{$O \in \mathbb{R}^{n_1 \times n_2}$} and pairs of matched  \textcolor{black}{columns $P \in \mathbb{R}^{min(n_1,n_2)}$}. Here, $n_1$ and $n_2$ represent the number of columns in the two schemas respectively.
\subsection{Schema Matching Components}

Table \ref{table:component} showcases six common SM types \cite{Koutras2020ValentineEM}. In this section, we will detail the key components of SMUTF: \textit{HXL-Style Tag Generation}, \textit{Rule-Based Feature Extraction} (which is subdivided into \textit{Column Name Features} and \textit{Value Features}), and \textit{Deep Embedding Feature Extraction}. In the context of previous SM or data discovery approaches \cite{nie2025snmatch,ma2025knowledge,seedat2025matchmaker,huang2024transform,liu2024gram,bogatu2020dataset, lehmberg2017stitching, yakout2012infogather, nargesian2018table, fernandez2018aurum}, the common practice is to employ 1 to 3 types of matchers. Differing from these traditional approaches, our proposed SMUTF system integrates five matchers, excluding the Value Overlap Matcher, as described in Table \ref{table:component} and illustrated in Figure \ref{fig:schema}. We consciously forgo the Value Overlap Matcher as its application is limited to a specific SM scenario: when two columns are joinable. Although this condition frequently occurs in practical situations, an excessive reliance on value overlap could unintentionally restrict the matching system's adaptability to various scenarios. A clear illustration of this limitation is the SM of event reports from different years (e.g., 2018 and 2019), where fields related to dates show no overlap. However, it's noteworthy that the distribution of values can to some extent substitute the role of value overlap. Therefore, SMUTF opts to use the Data Type Matcher and Distribution Matcher, which will be elaborated on in the subsequent sections. \textcolor{black}{The combination of different types of matchers guarantees that the proposed system is robust even if the given schemas are not complete (e.g. column names only or values only). This claim was justified in our ablation study shown in Section \ref{subsec:ablation}}.

The Humanitarian Exchange Language (HXL), a standard of tags developed to annotate the properties and content of column data, was originally proposed by the United Nations Office for the Coordination of Humanitarian Affairs \cite{kessler2015humanitarian}. This initiative aimed to enhance the efficiency and accuracy of data sharing related to humanitarian efforts. Currently, it is predominantly employed within the Humanitarian Data Exchange (HDX), a platform dedicated to the exchange of datasets.

Within SMUTF, we took the official HXL tag as a reference when creating HXL-style tags, utilizing them to increase interoperability and standardization among different datasets, which in turn refined our approach to SM.

HXL tags have two primary components: hashtags and attributes. Based on column data's content and formatting features, hashtags serve the purpose of delineating the primary categories of data, while attributes function as supplementary tags. A given column should possess only one hashtag, yet it may incorporate multiple attributes. For example, a column named ISO-3, comprising country codes such as USA, SSD, GBR, and so on, corresponds to an HXL tag set denoted as \textbf{"\#country+code+iso3"} where "country" is the hashtag and "code" and "iso3" are attributes.

In contrast to the original HXL tags, our HXL-style tags can include new hashtags to annotate data beyond the humanitarian field, providing a more flexible and extensible way of tagging data.

We employed Pre-trained Language Models (PLM), specifically GPT-4 \cite{openai2023gpt4} and mt0-xl \cite{sanh2022multitask}, which acted as teacher model and student model respectively, to automatically generate HXL-style tags. This approach captured the description of each column and its corresponding values to conduct a sequence-to-sequence task: essentially transforming one sequence of data (the raw data) into another (the tagged data).

One of the challenges we faced was the lack of datasets that came with pre-annotated HXL-style tags. We have tried using data from HDX, which includes HXL tags, as training data to train a model for generating HXL-style tags. However, the results were unsatisfactory. The primary reason is that HDX is a dataset from the humanitarian sector, where the data types and topics are too constrained. Once such a trained model is applied to open-domain datasets, it tends to generate erroneous HXL-style tags when encountering unfamiliar data.

To be specific, the HXL standard itself is designed for humanitarian workers, and its main hashtag categories are: (1) Places (2) Surveys and assessments (3) Responses and other operations (4) Cash and finance (5) Crises, incidents, and events (6) Metadata. Clearly, categories (2) to (5) are tailored for humanitarian assessments, disasters, and organizational operations. Unfortunately, data of other content types in reality are likely to all fall under the Metadata category (\#meta), leading to a serious limitation in the generated tags. Strictly speaking, HXL tags offer limited textual information for open-domain data processing, necessitating HXL-style tags that adhere to the basic HXL principles yet are capable of handling open-domain data.

To overcome the challenge, we used the concept of in-context learning \cite{min2022rethinking}, a machine learning approach where the model learns from the sequence of interactions during the dialogue, without being explicitly trained on a fixed dataset. \textcolor{black}{For in-context learning, such a sequence of interactions can be formatted into a few-shot prompt given to any generative model.}

\begin{figure}[h]
    \centering
    \includegraphics[width=\textwidth]{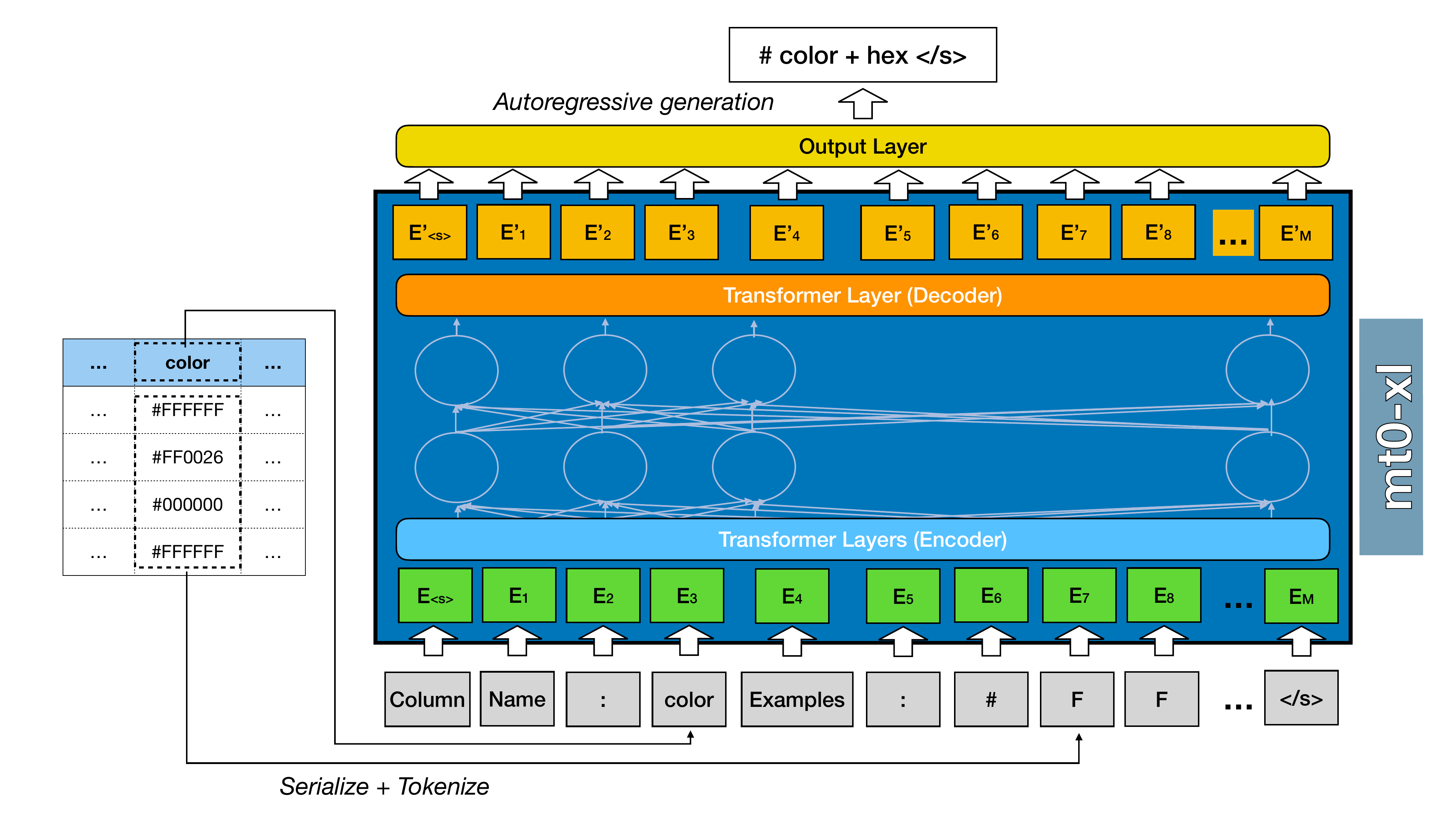}
    \caption{Generating HXL-style tags using mt0-xl model}
    \label{fig:hxltag_generate}
\end{figure}
\vspace{-0.15in}

\subsection{HXL-style Tags Generation}

In in-context learning, the initial step involved the formulation of training examples ${(x_i,y_i)}$ in a format that mapped inputs to labels using intuitive templates. \textcolor{black}{The $n$ training examples were integrated into a sequence as Equation \eqref{eq:0}:
\begin{align}
    P = \pi\{x_1, y_1\} \otimes \pi\{x_2, y_2\} \otimes \cdots \otimes \nonumber  \\
    \{x_{n}, y_{n}\} \otimes \pi\{x_{predict}, \ast \}
\label{eq:0}
\end{align}
where $\pi$ signifies a template-based transformation, and $\otimes$ represents the operation of concatenation. $x_{predict}$ is the input we want to label.
Also, the few-shot prompt helps generative models to understand our instruction to create tags}.

Below, we offer an example of a few-shot prompt used to generate HXL-style tags:

\lstset{
    basicstyle=\ttfamily\small, %
    breaklines=true, %
    xleftmargin=0pt, %
    xrightmargin=0pt, %
    breakindent=0pt
}

\begin{mdframed}
\begin{lstlisting}
I need help with predicting HXL-style tags, which annotate tabular data and consist of hashtags for primary categories and attributes for additional tagging, based on the data's content and format. Unlike standard HXL tags, HXL-style allows for creating new hashtags tailored to various topics, but the #meta hashtag should be avoided.

Each column is assigned a single hashtag and can have several attributes. For example, a column titled "ISO-3" containing country codes like "USA', "SSD', "GBR" would be tagged as "#country+code+iso3", where "#country" is the hashtag and "+code" and "+iso3" are its attributes. Hashtags begin with a # and attributes with a +.
\end{lstlisting}
\end{mdframed}

We provided five manually verified examples of HXL-style tag annotations, and the GPT-4 model learns to generate HXL-style tags in the context of these examples. We tested 1-shot, 5-shot and 10-shot examples within the prompt to generate HXL-style tags and asked two professional annotators, who both have a computer science-related master degree and own at least 3 years of academic experiences in natural language processing, to manually evaluate their generation qualities on 200 tag generation cases. Prompts with the 5-shot and 10-shot examples presented the best and similar generation accuracy, while the 5-shot generation was faster than the 10-shot generation. Some examples of HXL-style tags generated are listed in Table \ref{table:tag}. To evaluate the effectiveness of tag generation, we also conducted a manual review of the generated HXL-style tags. Two professional human annotators \textcolor{black}{with the background mentioned before} assessed a column sample (about 200 columns) and their corresponding HXL-style tags with aspects of Accuracy, Semantic Structuring, and Consistency. It is worth emphasizing that this was only evaluated to check the effectiveness of HXL-style tag generation results, and we did not make manual corrections, which would consume a lot of time and cost in practical applications.

Accuracy checks whether the tags correctly describe the associated data content or not. Semantic Structuring examines if the tags are appropriately structured in terms of class and attribute assignment, reflecting the intended semantics of the data. For the class signified by "\#", it should represent the primary category or the essence of the data point. For the attribute indicated by "+", it should function as a modifier to provide additional context or specificity to the class. Consistency means that the same data or concept should be consistently represented by the same tags across the whole sample set.

Each human annotator made a binary judgment on each data point, deciding whether the corresponding HXL-style tag was acceptable with respect to the given aspect. In terms of assessment results, the average acceptability for Accuracy, Semantic Structuring, and Consistency were 98.37\%, 95.38\%, and 89.67\% respectively. The inter-annotator reliability (Cohen's kappa coefficient) for the three aspects was 0.66, 0.69, and 0.59, respectively. This indicates that the tags generated by the GPT-4 methodology exhibited relatively good performance. The lower acceptability for Consistency was primarily due to the dominant effect of column names on GPT-4, leading to inconsistent granularity judgments. For example, for a column named "media\_thumbnail" that contains image links, the generated result was "\#media+thumbnail", but a similar column with a meaningless name like "col3" yielded the corresponding result of "\#url+image".

\begin{table}[htb]
  \centering
  \caption{HXL-style Tag Examples}
  \resizebox{\columnwidth}{!}{%
      \begin{tabular}{ccc}
        \toprule
        Column Name & Column Values & Generated HXL-style Tag \\
        \midrule
        color & \#FF0026, \#FFFFFF, \#FFFFFF & \textbf{\#color+hex} \\
        article\_url & https://www.dcard.tw/... & \textbf{\#url+article} \\
        like & 4123, 281842, 13 & \textbf{\#like+count} \\
        col4 & \$199, \$91, \$66 & \textbf{\#value+usd} \\
        \bottomrule
      \end{tabular}
  }
  \label{table:tag}
  \vspace{-0.1in}
\end{table}

Next, we used the generated tags to fine-tune the mt0-xl model \cite{muennighoff2022crosslingual}, a sequence-to-sequence text generation model with 3.7 billion parameters. We used the parameter-efficient fine-tuning methodology, LoRA \cite{hu2021lora}, to adapt the model to our task of generating HXL-style tags. This helps us create a powerful model capable of understanding and generating HXL-style tags effectively within the computational constraints of our GPU capacity. The generation process is shown in Figure \ref{fig:hxltag_generate}.

In SMUTF, We calculated the matching score of HXL tags between the source and target columns. We computed $\mathbf{h_{i,j}}$, which is a concatenation of the exact match score ($\mathrm{E_{tag}}{\mathit{i},\mathit{j}}$) on hashtag and the Jaccard similarity score ($\mathrm{jac_{tag}}_{\mathit{i},\mathit{j}}$) on attributes, to reflect this metric. Equation \ref{eq:3} provides the formulas used for calculating the HXL tags matching score.

\begin{equation}
    \mathbf{h_{i,j}} = \left[ \mathrm{E_{tag}}_{\mathit{i},\mathit{j}};\mathrm{jac_{tag}}_{\mathit{i},\mathit{j}} \right]
    \label{eq:3}
\end{equation}
\vspace{-0.1in}
\subsection{Rule-based Feature Extraction}

To apply machine learning on SM has been studied in various works \cite{sahay2020schema, berlin2002database}. The initial step in this process requires feature engineering, which involves the extraction of distinct descriptors that represent the attributes of column names and values. The goal is to facilitate comparison and differentiation between columns, thereby providing an estimate of their similarity or dissimilarity. Features can be classified based on columns' textual representation, especially the column names; they can also be categorized according to the values that columns correspond to, since features for numerical data, dates, and text would differ significantly. In this section, we describe the features we used, including those derived from column names and those derived from values.

\paragraph{\textbf{Column Name Features}}

The Column Name Features were calculated through pairwise comparisons between columns, where we employed various metrics to measure string similarity.

\begin{itemize}
    \item \textbf{BLEU Score:} This metric computes the Bilingual Evaluation Understudy score \cite{papineni2002bleu} between two column names. Given that BLEU is traditionally used for evaluating the similarity between machine-translated and human-translated texts, it can effectively measure the similarity between column names, especially when considering semantic nuances.
    \item \textbf{Edit Distance:} This metric computes the Damerau Levenshtein distance \cite{levenshtein1966binary}, which measures the edit distance between two strings with substitutions, insertions, deletions, and transpositions. The Damerau-Levenshtein distance acknowledges the human tendency to make certain typos \cite{mawardi2020spelling}, such as transpositions, making it a versatile measure for comparing column names.
    \item \textbf{Longest Common Subsequence Ratio:} This metric represents the Longest Common Subsequence ratio between two column names. This helps gauge how many continuous letters of one column name appears in another, which can be a potent signal when the columns have long descriptive names.
    \item \textbf{One-In-One Occurrence:} The feature $o_{\mathit{i,j}}$ is a binary indicator demonstrating whether the name of one column is included within another. Here, $c_i$ and $c_j$ refer to the names of the columns. The presence of one column name within another can indicate a sub-category or related attribute.
    \begin{equation}
    o_{\mathit{i,j}} = 
    \begin{cases}
    1 & \text{if } c_i \in c_j \vee c_j \in c_i, \\
    0 & \text{otherwise}.
    \end{cases}
    \label{eq:one}
    \end{equation}
    \item \textbf{Cosine Similarity in Semantic Embedding} 
    The similarity is calculated as the cosine similarity score \cite{7577578} between the semantic embeddings, $\mathbf{s_i}$ and $\mathbf{s_j}$, of two column names. Details of the semantic embedding process will be expounded in the following section. This is especially useful when names might not be lexically similar, but they convey related concepts.
\end{itemize}

Consequently, we derive the formula \ref{eq:2} to compute the feature score $\mathbf{ls_{i, j}}$ of the column names $c_{\mathit{i}}$ and $c_{\mathit{j}}$.

\begin{equation}
    \mathbf{ls_{i,j}} = \left[ \frac{\mathbf{s_{i}}\cdot \mathbf{s_{j}}}{\left | \mathbf{s_{i}} \right | \left | \mathbf{s_{j}} \right |};\mathrm{bleu}(c_{\mathit{i}},c_{\mathit{j}});\mathrm{lev}(c_{\mathit{i}},c_{\mathit{j}});\mathrm{lcs}(c_{\mathit{i}},c_{\mathit{j}}); o_{\mathit{i,j}} \right]
    \label{eq:2}
\end{equation}

In this formula, $i$ and $j$ represent the indices of the source column and the target column, respectively. $\mathrm{lev}$ and $\mathrm{lcs}$ refer to the edit distance and longest common sub-sequence ratio respectively.

\paragraph{\textbf{Value Features}}

The value features were derived by analyzing the characteristics of the values, such as data type and numerical distribution. As they represented the distribution or type features of individual columns, they could not explicitly reflect the similarity between the values of two columns. To address this, we introduced a normalization formula to calculate the similarity score between the value features of two columns $i$ and $j$:

\begin{equation}
  \mathbf{lv_{i,j}} = \frac{{\left| \mathbf{f_i} - \mathbf{f_j} \right|}}{{\mathbf{f_i} + \mathbf{f_j} + \epsilon}} 
\label{eq:vs}
\end{equation}

In this formula, $\mathbf{f_i}$ and $\mathbf{f_j}$ refer to the computed value features of the i-th and j-th columns, respectively. $\epsilon$ is an error term of small number in order to avoid a zero denominator. Here, the division is element-wise division. The resulting score captures the relative difference between the two sets of column value features, effectively serving as a similarity measure. This score is aptly suited for subsequent learning using the gradient boosting algorithm.

\begin{itemize}
\item \textbf{Data Type Features:} These are applicable to all types of data, and make use of one-hot encoding to convert categorical data to a binary format. By identifying the inherent type of the data, we can have an initial grasp of what kind of information the column might be conveying, and this can also aid downstream processes that might handle different types in different ways. They include: 

    \begin{itemize}
    \item \textbf{URL Indicator:} A binary feature indicating whether the data is a URL.
    \item \textbf{Numeric Indicator:} A binary feature indicating whether the data is numeric.
    \item \textbf{Date Indicator:} A binary feature indicating whether the data is a date.
    \item \textbf{String Indicator:} A binary feature indicating whether the data is a string.
    \end{itemize}
    
\item \textbf{Length Features:} These features give an overview of the variety and complexity of the data, applicable to all types. These give a snapshot of the richness and diversity of the data. For example, a column with a high variance in length might indicate free-text inputs, while one with a low variance could indicate fixed-form data.

    \begin{itemize}
    \item \textbf{Mean, Minimum, Maximum Length:} Indicate the central tendency and range of data string lengths.
    \item \textbf{Variance and Coefficient of Variation (CV):} Measure the diversity and consistency of the data string lengths.
    \item \textbf{Unique Length to Data Length Ratio:} Quantify the richness and uniqueness of data string lengths.
    \end{itemize}

\item \textbf{Numerical Features:} These features focus on the numerical aspects of the data, applicable only to numerical data. By understanding the distribution and characteristics of numerical data, we can make initial assessments about the nature of the column – for instance, a column with a unique-to-length ratio near 1 might indicate unique identifiers.

    \begin{itemize}
    \item \textbf{Mean, Minimum, Maximum:} Describe the central tendency and range of the numerical values.
    \item \textbf{Variance and CV:} Assess the variability and relative dispersion of the numerical values.
    \item \textbf{Unique to Length Ratio:} Compute the ratio of unique values to the total number of values, reflecting the tendency of potential outliers over common values.
    \end{itemize}

\item \textbf{Text Features:} These features analyze the structure and semantics of non-numerical data. The textual nature of data holds valuable information. Understanding patterns in whitespace, punctuation, and other character types can hint at the structure, composition, and complexity of the data. 

    \begin{itemize}
    \item \textbf{Mean and CV of Whitespace, Punctuation, Special Character, and Numeric Ratios:} Compute the mean and CV of the ratio of different special characters in the data string. In the ablation experiments, this part of the text features that do not involve semantics is called character features. %
    \item \textbf{Semantic Embedding:} Reveal the underlying contextual meanings and relationships between data strings, computed as the average embedding of 20 randomly selected textual values.
    \end{itemize}
\end{itemize}
\vspace{-0.1in}
\subsection{Deep Embedding Similarity}

Every column, either from the source or the target column set, was transformed into deep embeddings. These consisted of a column name embedding, $\mathbf{s}$, and a textual value embedding, $\mathbf{t}$ (only for columns with text features). For each column name and textual value set, we employed a fine-tuned multilingual pre-trained language model, MPNet \cite{Reimers2019SentenceBERTSE,reimers-gurevych-2020-making, song2020mpnet}, to construct semantic embeddings. This involved tokenizing each column name and value set and then passing them through the model individually. The embeddings for an entire column were computed by aggregating the output of the model for each token. As validated by Table \ref{tab:sentence_embedding}, the multilingual MPNet displayed superior performance on a variety of sentence-pair tasks, especially semantic textual similarity, aligning well with our objective of assessing the similarity between two deep embeddings of column names.
\vspace{-0.1in}
\subsection{Similarity Score Prediction using XGBoost}

An ultimate hybrid similarity feature $\mathbf{l_{i,j}}$ is obtained from the $\mathbf{ls_{i,j}}$ (see Eq. \ref{eq:2}), $\mathbf{h_{i,j}}$ (see Eq. \ref{eq:3}), cosine similarity of textual value embeddings $\frac{\mathbf{t_{i}}\cdot \mathbf{t_{j}}}{\left | \mathbf{t_{i}} \right | \left | \mathbf{t_{j}} \right |}$ and value feature score $\mathbf{lv_{i,j}}$ (see Eq. \ref{eq:vs}). A classifier takes $\mathbf{l_{i,j}}$ as an input to predict if $c_i$ and $c_j$ are matched. 

\begin{equation}
    \mathbf{l_{i,j}} = \left[ \mathbf{ls_{i,j}};\mathbf{lv_{i,j}};\frac{\mathbf{t_{i}}\cdot \mathbf{t_{j}}}{\left | \mathbf{t_{i}} \right | \left | \mathbf{t_{j}} \right |}; \mathbf{h_{i,j}}  \right] \label{eq:1}
\end{equation}

\textcolor{black}{The similarity score prediction was framed as a binary classification task, and the SM system was not bounded with any specific machine learning model; instead, every existing binary classification model could be deployed here.} We used an XGBoost classifier, a scalable and high-performing tree boosting system, to predict a matched pair given the hybrid similarity feature. \textcolor{black}{The motivation to use XGBoost is that it delivers a more impressive prediction on SM than other machine learning techniques like neural networks or LightGBM, as shown in Table \ref{tab:ml_model}, where other models' predictions are evaluated.}

The output of the XGBoost models was a match score, indicating the probability that two columns are matched. While default thresholds were computed, users could define custom thresholds for deciding a match. The default threshold of SMUTF is chosen based on the best performance on the evaluation portion of the training dataset, so it is influenced by the training dataset. The threshold typically falls within the range of 0.1 to 0.15, although this value may vary depending on the feature selection and the specific training data used.

\textcolor{black}{We aimed to enhance the robustness of our model training by employing a multi-model training strategy. Our data was partitioned into 16 subsets, where each subset was used as a validation set once, with the remaining 15 subsets serving as the training set during that iteration. This partitioning resulted in the creation of 16 distinct XGBoost models, each with its own trained weights and hyper-parameters determined by its assigned training and validation set.}

\textcolor{black}{To consolidate the predictions from all 16 models, we applied a soft voting fusion mechanism. The majority vote from these models was then used as the final matching decision. Additionally, for the composite similarity score, we computed the average of the scores generated by all individual models.}

\textcolor{black}{This approach not only mitigates data bias caused by dataset partitioning but also enhances the stability of predictions, leading to a more reliable similarity score. By integrating deep embedding similarity with the XGBoost-based similarity score prediction, our method effectively supports multilingual semantic similarities and provides adaptability with custom thresholds.}

\section{Datasets}

In this section, we will introduce the training dataset used for SMUTF, the proposed HDXSM Dataset used for schema matching (SM) evaluation, and other publicly available evaluation datasets proposed by previous research studies.

\begin{table}[htp]
\centering
\caption{Statistics of Training Dataset}
\small %
\renewcommand{\arraystretch}{0.7}
\resizebox{\textwidth}{!}{%
\begin{tabular} {lllrrlll}
\toprule
No. & Data Source 1\tnote{a} & Data Source 2\tnote{b} & \# Column & \# \revise{Matched Col Pairs} & Language(Columns) & Language(Values) & Theme\\
\midrule
1 & Sinyi Realty & Eastern Realty & 15, 14 & 12 & Chinese & Chinese & Real Estate\\
2 & Atmovies & Yahoo Kimo Movies & 4, 6 & 3 & Chinese & Chinese, English & Movie\\
3 & Bahamut Forum & PTT-C\_Chat & 4, 4 & 4 & Chinese, English & Chinese & Anime\\
4 & RT-Mart & Carrefour & 5, 4 & 4 & Chinese, English & Chinese & Retailers\\
5 & TVBS News & UDN News & 4, 5 & 4 & Chinese & Chinese & News\\
6 & Liberty Times & Reddit & 5, 6 & 5 & English & English, Chinese & News Forum\\
7 & Bahamut Animation & bilibili & 6, 7 & 5 & English & Chinese & Danmaku\\
8 & Baidu Baijia & Wanxiang Patent & 4, 5 & 4 & Chinese & Chinese, English & cosmetic\\
9 & PChome & Shopee & 3, 4 & 3 & English & Chinese & E-commerce\\
10 & kkday & klook & 9, 9 & 8 & English & Chinese & Tourism\\
11 & PChome & MOMOshop & 7, 5 & 5 & English & Chinese & E-commerce\\
12 & cosme & wastsons & 5, 5 & 5 & Chinese & Chinese & cosmetic\\
13 & Dcard-part time & PTT-part time & 4, 5 & 4 & English & Chinese & Job\\
14 & Webtoon & kuaikanmanhua & 7, 6 & 5 & English & Chinese & Comics\\
15 & Dcard & PTT Forum & 4, 6 & 4 & English & Chinese & Politics\\
16 & IMDb & Rotten Tomatoes & 3, 4 & 3 & English & English & Movie\\
17 & Scopus-articles & Scopus-author & 7, 8 & 3 & English & English & Academic\\
18 & TPC-H & TPC-H & 6, 6 & 6 & English & English & Customer\\
19 & Altermidya & Pagina12 & 10, 10 & 10 & English & English, Spanish & News \\
\bottomrule
\end{tabular}
}
\begin{tablenotes}
  \item[a,b] Each table includes a minimum of 20 values.
  \item Only the first 16 pairs were used for training the model, while the remaining data was solely used for testing and evaluation.
\end{tablenotes}
\label{table:training_data}
\vspace{-0.1in}
\end{table}
\vspace{-0.1in}
\subsection{Training Dataset}

\renewcommand{\arraystretch}{0.9} %

Table \ref{table:training_data} presents the sources of our training dataset, the column count for each table, along with the corresponding languages and topics, etc. We primarily obtained data from various popular websites through web scraping. Since the data was publicly available on the internet, the content was diverse in themes, including movies, real estate, animation, online shopping, cosmetics, and more. To create ground truths of the training dataset, we explored every pair of tables belonging to the same theme, and then columns with potential matches were subjected to manual alignment. This manual alignment process was undertaken by a team of four human annotators who deliberated over each table pair individually, reaching consensus before finalizing their decisions. During this process, care was taken to ensure that each annotator comprehended the content of the websites, as well as the data within the table pairs, ensuring a unanimous agreement was reached without dispute.

It's important to note that none of the topics in the training data appeared in the evaluation data. The only exception to this was pair number 16, where there were some overlaps between the IMDb website in our training data and the MovieLens-IMDB in the evaluation data. However, even though both involved the IMDb website, the columns used were different. The training dataset used a minority of columns related to movie ratings from IMDb, while MovieLens-IMDB involved aspects such as movie classification, theme, etc., which were not present in the training data.

Simultaneously, the volatility of online data conferred pattern matching value on these contents. For example, the same scoring metric might have been named "rating\_star" on website A, while website B might have named it "review\_star". In addition, to increase the complexity of pattern matching and prevent the model from simply deducing inter-column relationships via column names, we manually modified certain columns. The main modification methods included language translation or masking. Language translation involved converting original Chinese column names into English, while masking changed meaningful column names into nonsensical codenames like "col3".

Ultimately, the majority of websites we collected featured content in either Chinese or a mixture of Chinese and English. This implicated the multilingual embedding component within SMUTF. Multilingual data could enhance the model's robustness when facing datasets from other domains. Moreover, if our system had been trained primarily on Chinese datasets and could demonstrate effective results in English domain data without additional training or fine-tuning, it would provide further evidence of our system's performance in an open-domain scenario.

To enrich the complexity of the data, our dataset not only contained text-based information but also a significant volume of identifiers, values, dates, URLs, and other forms of data. Such diversity of data types is common in pattern-matching tasks. By integrating different data forms, our model was expected to capture more complex patterns and achieve better, demonstrating a broader applicability to a variety of real-world data scenarios.
\vspace{-0.1in}
\subsection{HDXSM Dataset}

Although SM has been an established research area for decades, it always suffers from a lack of publicly available, large-scale, real-world datasets. Current studies have predominantly relied on datasets that are automatically generated based on specific rules, or they have employed small-scale real-world datasets for method evaluation. Furthermore, most high-quality, real-world evaluation datasets from industry may not be accessible to the public due to privacy concerns. Recognizing these limitations, we developed a larger-scale, real-world SM dataset. This new dataset, named HDXSM, used data from the Humanitarian Data Exchange (HDX) and was annotated with existing HXL tags and extensive manual checks.

As of May 2023, the HDX had amassed a repository of 20,881 datasets, of which 8,652 had been adorned with HXL tags. Our research specifically focused on data delineated in tabular form; hence, we confined our analysis to datasets in CSV, XLS, and XLSX formats, which involves 8,640 datasets. It is important to note that each dataset may comprise multiple tables.

In line with the premise that HXL provides an accurate representation of column names and value data, we posited that two columns featuring identical HXL tags (including both hashtags and attributes) were eligible for matching. This led us to the inherent challenge of dataset selection for SM. The objective of our methodology was to faithfully replicate or mirror the practical requirements of humanitarian workers. Given this context, random pairwise matching of datasets was often impractical. For instance, cross-matching a food price dataset from Zimbabwe with a population dataset from Vietnam was devoid of tangible significance. Humanitarian work generally entails long-term commitment in specific regions, which necessitates the frequent linkage and analysis of data within a confined area (e.g., a particular country). Additionally, the data slated for linkage should originate from identical or overlapping domains. Fortunately, HDX provides a wealth of metadata for each dataset. Our methodology chiefly harnessed "groups" (\textcolor{black}{indicating} the countries involved in the dataset) and "theme tags" (representing the thematic or domain-specific aspects addressed in the dataset, such as COVID-19, funding, etc.).
\vspace{-0.001in}
During the assembly of the HDXSM dataset, our process initially involved traversing all datasets for each country. A pair of datasets was deemed suitable for matching if both pertained to the same country and their theme tags yielded a Jaccard similarity exceeding 0.4. Subsequently, all tables from the two datasets were extracted and the HXL tags of each column were juxtaposed. However, certain datasets exhibited a high degree of similarity, potentially attributed only to differing data collection timelines, especially among regular observational datasets. These datasets resulted in an abundance of duplicate matches. We circumvented this redundancy by discarding repetitive matches, retaining only those pairings that showcased unique attributes in terms of column name and HXL tags. A subsequent review of the data revealed values of erroneous matching. These inaccuracies, unrelated to our methodology, were ascribed to pre-existing HXL tags annotation errors within the HDX datasets, such as values of reverse annotation of HXL tags for two columns. Therefore, we instigated a comprehensive manual annotation check of the entire HDXSM dataset. Ultimately, the HDXSM dataset incorporated a total of 204 table pairs. Each table contained a maximum of 100 rows, with the aggregate of columns across all tables amounting to 9,394. Notably, out of these, 2,635 column pairs were matched.
\vspace{-0.1in}
\subsection{Publicly Available Datasets}

We incorporated four public available datasets into our experiment. The \textbf{WikiData} dataset comed from \textit{Valentine}\footnote{https://delftdata.github.io/valentine/} \cite{Koutras2020ValentineEM}. It was collected from real-world data and re-organized into different types of schema pairs. Given a tabular schema, \textit{Valentine} suggests splitting it horizontally to create \textit{unionable} pairs, vertically to make \textit{joinable} pairs, or in both ways. This methodology helps resolve the limited data sources for SM by manually generating new column pairs from a single table. To be specific, a \textit{unionable} dataset is created by horizontally partitioning the table with different percentages of row overlap, and a \textit{view-unionable} dataset is made by splitting a table both horizontally and vertically with no row overlap but various column overlap. A pair of \textit{joinable} tables should have at least one column in common and a large row overlap. The \textit{semantically-joinable} dataset is similar to the \textit{joinable} one, except that their column names are noisy (semantic variations). In addition, all values under different types of datasets are manually made noisy. WikiData has 4 schema pairs, and each table has 13-20 columns. The maximum row number can be above 10,000.

The second publicly available dataset was obtained from two public movie databases, MovieLens\footnote{https://grouplens.org/datasets/movielens/} and IMDB\footnote{https://www.imdb.com/interfaces/}. These two databases are commonly used to create schema pairs since their columns are similar to each other, like \textit{rating} vs. \textit{averageRating} or \textit{title} vs. \textit{originalTitle}. The MovieLens-IMDB dataset has been widely used in the field of SM \cite{zhang2023schema, Zhang2011AutomaticDO}, but there is not a standard version of it. Our \textbf{MovieLens-IMDB} dataset has 2 pairs of schemas and each schema has 1000 rows. Its column number varies from 4 to 10. 

The final pair of datasets, \textbf{Monitor} and \textbf{Camera}, originate from the \textit{DI2KG Benchmark} datasets\footnote{http://di2kg.inf.uniroma3.it/datasets.html}. DI2KG is acknowledged as a comprehensive data integration benchmark that comes with a mediated schema. These datasets encompass product specifications scraped from a wide range of eCommerce platforms such as ebay and walmart. We utilize their mediated schema for precise schema matching, linking source attributes (e.g., "producer name" from eBay) to target attributes (e.g., "brand"). This process ensures data consistency across diverse eCommerce platforms by matching attributes under a closed-world assumption, exemplified in our creation of 20 table pairs with detailed matches. This meticulous alignment allows for structured data integration, facilitating comparisons and analysis within our research. A distinctive attribute of the Monitor and Camera datasets is the prevalence of numerous many-to-many correspondences; here, a unique column may find matches across multiple columns within a disparate table, an intricacy brought forth by the application of the mediated schema.

The basic statistics of all the benchmark datasets including HDXSM are given in Table \ref{table:datasets}.

\begin{table}[h]
\caption{Statistics of Benchmark Datasets}
\resizebox{\columnwidth}{!}{%
\begin{tblr}{lrrrr}
\toprule
Dataset & \SetCell[c=1]{c} \# Table Pairs & \SetCell[c=1]{c} Avg \# Rows & \SetCell[c=1]{c} \# Cols & \SetCell[c=1]{c} \# Matched Cols \\ 
\midrule
WikiData & 4 & 9489 & 120 & 40 \\
MovieLens-IMDB & 2 & 1000 & 23 & 5 \\
Monitor & 20 & 406 & 1582 & 584 \\ 
Camera & 20 & 793 & 1465 & 567 \\ \hline
HDXSM & 204 & 100 & 9394 & 2635 \\
\bottomrule
\end{tblr}%
}
\label{table:datasets}
\vspace{-0.1in}
\end{table}

\section{Experiments}

We evaluated the performance of SMUTF across four distinct datasets and six benchmark approaches. Our evaluation utilizes macro-F1 and macro-AUC scores to compare the performance of our method with the benchmarks.
\vspace{-0.1in}
\subsection{Evaluation Metrics}

 Every dataset is made up of schema pairs, and each of them represents \revise{an} SM task to be solved. Initially, we determine an F1 and an AUC score of ROC for each schema pair. Following that, we compute the average F1 and AUC score across all the schema pairs contained in the dataset. We refer to these average scores as the macro metrics of the dataset (macro-F1 and macro-AUC).
\vspace{-0.1in}
\subsection{Benchmarking Methods}

Benchmarking SM approaches evaluated in experiments can be tentatively categorized into  three types: schema-based, value-based and hybrid matching.

\subsubsection{Schema-based Matching} 

A schema-based matching employs schema-related information, which includes column names, description, inter-column relationships, etc, to find out matched pairs within two different schemas. 

\paragraph{\textbf{Cupid}} The Cupid \cite{madhavan2001generic} framework represents an initial  effort of this approach, encompassing linguistic matching of column names, which calculates similarity through synonyms and hypernyms, and structural matching that examines the hierarchy between columns, considering their containment relationships. Column matches are determined by a weighted combination of these linguistic and structural similarities.

\paragraph{\textbf{Similarity Flooding}} Similarity flooding \cite{melnik2002similarity} is a schema  matching method that uses graph representations to assess relationships between columns, initiating with a string matcher that identifies potential column matches through common prefixes and suffixes. This method then expands the search for matches by propagating similarities; if two columns from different schemas are similar, their neighboring columns' similarity is also increased. Like Cupid, similarity flooding heavily relies on the linguistic resemblance of column names.

\paragraph{\textbf{COMA}} COMA \cite{Do2002COMAA} introduces a system that flexibly integrates multiple schema matchers to evaluate column similarity across different schemas. Schemas are modeled as rooted directed acyclic graphs, with each column represented by a path from the root. COMA employs various strategies to aggregate the similarity scores provided by different matchers, such as taking the average or maximum, and it uses specific criteria to select matching column pairs, like those exceeding a similarity threshold or ranking in the top-\textit{K}. Experimental results indicate that while individual matchers might be flawed, their combined use can enhance matching performance. COMA has evolved to include instance-based matching, leading to two variants: \textbf{COMA-Schema} for schema-centric matching and \textbf{COMA-Instance}, which adopts a hybrid matching approach.

\subsubsection{Value-based Matching} 

A value-based matching is data-oriented, focusing on utilizing statistical measures to explore relationships between values under different columns. 

\paragraph{\textbf{Distribution-based}} A distribution-based schema matcher \cite{Zhang2011AutomaticDO}  utilizes the Earth Mover's Distance (EMD) to measure how much effort is required to transform one column's set of values into another, focusing on their rankings. Initially, it clusters columns using pairwise EMD calculations. Next, clusters are broken down into matched pairs using the intersection EMD, a metric grounded in two principles: columns sharing many values are likely related, and columns with minimal intersection are matched if they both significantly overlap with a third column.

\paragraph{\textbf{Jaccard-Levenshtein}} The Jaccard-Levenshtein method \cite{Koutras2020ValentineEM} is a value-based SM technique that applies the Jaccard similarity index to evaluate the relatedness between pairs of columns, considering two values as identical if their Levenshtein distance falls below a predefined threshold. This approach offers a direct and uncomplicated way to compare the distribution of values in columns to ascertain matches.

\subsubsection{Hybrid Matching}

A hybrid SM involving the consideration of both schema-related information and value-oriented features. \textit{\textbf{EmbDI}} \cite{cappuzzo2020creating} is an approach for data integration by creating relational embeddings upon column names and values. These embeddings are trained from scratch and external knowledge such as synonym dictionaries is involved. Our proposed model, SMUTF, is also an hybrid matching-based model, since it not only builds semantic embeddings on column names but also compute value features when comparing two columns. The integration of schema-based information and  column values is supposed to present a more robust performance upon SM than matching techniques with a single focus. 

\begin{table}[!htp]
  \centering
  \caption{The effectiveness of SMUTF is evaluated against other benchmarks. The metrics employed for assessment in the experiment are the macro-F1 and AUC. The one in \textbf{bold} is the top-ranked result, while the result underlined comes in as the second best.}
  \vspace{0.01in}
  \resizebox{\textwidth}{!}{%
  \begin{tabular}{@{}llrrrrr@{}}
    \toprule
    \multirow{3}{*}{Method}  & \multicolumn{6}{c}{Datasets (F1 / AUC)} \\
    \cmidrule{3-7}
    & Avg Score &  WikiData & MovieLens-IMDB & Monitor & Camera & HDXSM \\  
    \midrule
     \multirow{1}{*}{Cupid \cite{madhavan2001generic}} & 58.27 / 85.83  & 66.76 / 88.52  & 83.33 / \underline{98.65} & 34.31 / 75.97 & 35.22 / 77.90 & 71.73 / 88.11 \\ 
    \multirow{1}{*}{Similar Flood \cite{melnik2002similarity}} & 53.73 / 90.62 & 68.89 / 92.25 & 61.91 / 98.20 & 32.32 / 82.57 & 35.26 / 86.96 & 70.27 / 93.10 \\
    \multirow{1}{*}{COMA-Schema \cite{Do2002COMAA}} & 57.71	/ 80.60  &  67.18 / 85.89 & 83.33 / 90.99 & 35.50 / 65.58 & 39.82 / 72.68 & 62.70 / 87.88  \\ 
    \multirow{1}{*}{COMA-Instance \cite{Do2002COMAA}} & 62.56	/ 84.57  &  88.10 / 98.13 & 74.07 / 91.44 & 43.91 / 68.94 & 42.60 / 73.32 & 64.13 / 91.01  \\ 
    \multirow{1}{*}{Distribution \cite{Zhang2011AutomaticDO}} & 44.67 / 72.20 &  72.80 / 89.71 & \underline{90.00} / 89.60 & 15.22 / 60.04 & 10.80 / 54.14 & 34.51 / 67.53 \\
    \multirow{1}{*}{EmbDI \cite{cappuzzo2020creating}} & 46.33 / 74.22  &  72.62 / 93.88 & 68.57 / 81.37 & 26.93 / 63.41 & 21.53 / 60.11 & 41.98 / 72.32 \\
    \multirow{1}{*}{Jaccard Leven \cite{Koutras2020ValentineEM}} & 56.42	/ 81.27 &  \underline{89.27} / 96.90 & 80.00 / 92.23 & 30.05 / 69.19 & 30.16 / 63.39 & 52.61 / 84.63 \\ 
    \multirow{1}{*}{Sherlock \cite{hulsebos2019sherlock}} & 31.73	/ 61.35 &  56.45 / 70.36 & 53.57 / 54.16 & 7.02 / 51.65 & 7.11 / 57.89 & 34.51 / 72.71 \\ 
    \midrule
    \multirow{1}{*}{SMUTF w/o tag} & \underline{71.64} / \underline{95.65} &  85.47 / \underline{98.78} & $\textbf{100.00}$ / $\textbf{100.00}$ & \underline{44.03} / \textbf{89.88} & \underline{49.80} / \textbf{92.08} & \underline{78.92} / \underline{97.52} \\ 
    \multirow{1}{*}{SMUTF} &  \textbf{74.40} / \textbf{95.70} & $\textbf{91.39}$ / $\textbf{99.53}$ & $\textbf{100.00}$ / $\textbf{100.00}$ & \textbf{45.15} / \underline{89.44} & \textbf{52.40} / \underline{91.32} & \textbf{83.04} / \textbf{98.20} \\
    \bottomrule
  \end{tabular}
}
  \label{table:comparison}
  \vspace{-0.1in}
\end{table}

\begin{figure*}[h]
    \centering
    \includegraphics[width=\textwidth]{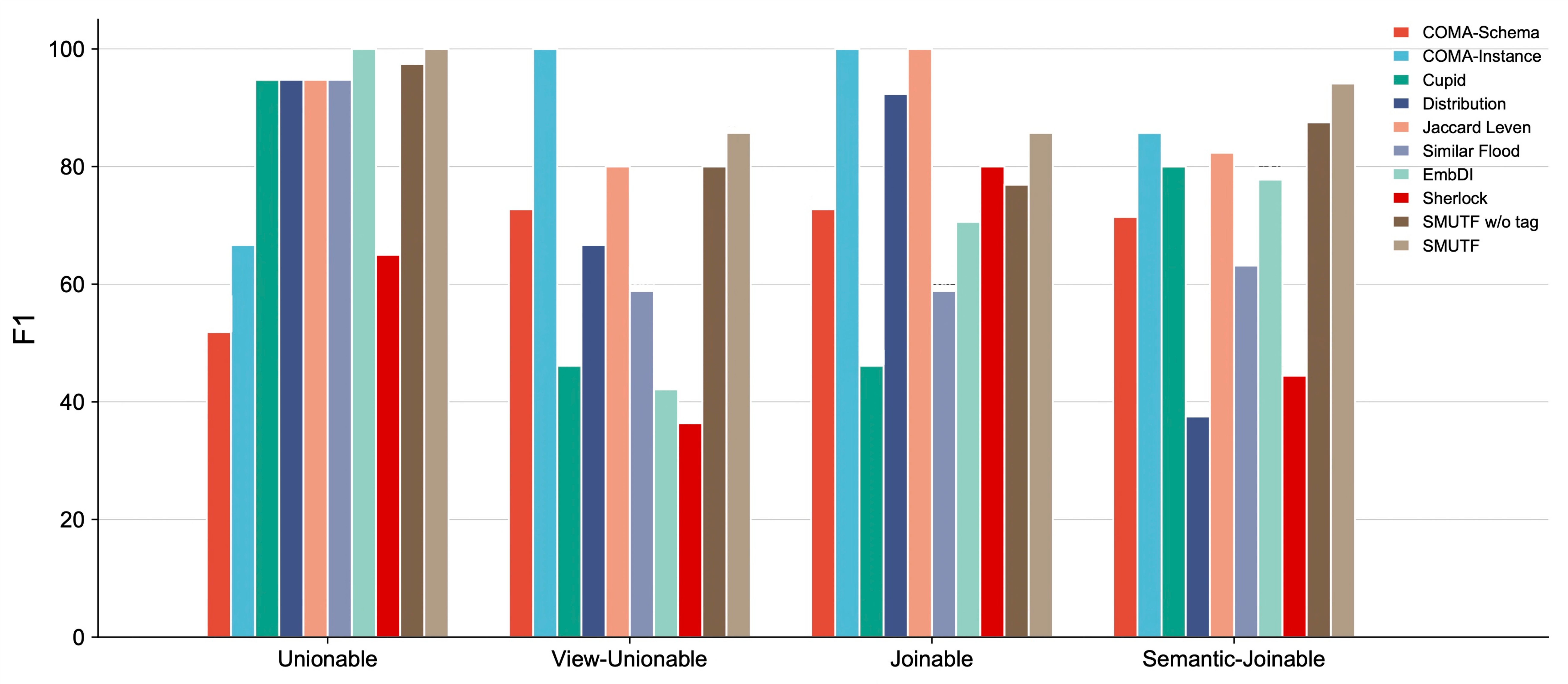}
    \caption{Performance of different methods upon different schema pairs of WikiData is explored. The metrics employed for assessment in the experiment is the F1.}
    \label{fig:wikidata}
    \vspace{-0.1in}
\end{figure*}

Furthermore, the SMUTF framework's generation of HXL-style tags can be regarded as a  semantic annotation technique identifying data types, which enriches the landscape of SM strategies. Within this context, we have incorporated \textit{\textbf{Sherlock}} \cite{hulsebos2019sherlock} into our suite of benchmark methods. Sherlock operates through an elaborate supervised learning paradigm, processing an extensive corpus of tabular datasets. It adeptly derives a diverse array of features from both columnar nomenclature and cell contents, subsequently assigning these to  different types of semantic data, such as Location, Name, or Year. In the application within the SM domain, we match columns that Sherlock identifies as being of the same data type.

\vspace{-0.1in}
\subsection{Benchmark Results and Discussion}

The inference results for various datasets were displayed in Table \ref{table:comparison} and Figure \ref{fig:wikidata}. To assess the versatility of our model across different domains,  we ensured that the domains of inference datasets differed from those of the training dataset.

Wikidata provided by \textit{valentine} had different types of datasets based on four table-splitting strategies (see Figure \ref{fig:wikidata}). Though our model's macro-F1 performance was the best one among other benchmarks (see Figure \ref{fig:wikidata}), its individual evaluation on the \textit{joinable} dataset (F1 85.71\%) was worse than the Jaccard-Levenshtein method (F1 100\%). In a \textit{joinable} pair of schemas, there was a large overlap of rows. Since the Jaccard-Levenshtein method is a naive approach matching columns based on the row-value distribution from each schema, it was not surprising that given a large number of rows in each schema (more than 5000), this model could do a perfect SM. This observation was also found in the performance comparison between schema-based (COMA, Cupid, similarity flooding) and value-based (distribution-based, Jaccard-Levenshtein) methods, where data-oriented methods did a better job on the WikiData than the schema-based methods. In addition, as mentioned in the Method section, we did not include a value overlap matcher in SMUTF. This was also the reason for the poorer performance.

Compared to the Jaccard-Levenshtein method that purely focuses on values, SMUTF considered the semantic variations of column names. As a result, in the \textit{sem-joinable} dataset, the Jaccard-Levenshtein method's F1 score (82.35\%) was less than SMUTF's (94.12\%). We also noticed that without the addition of HXL-style tags (see Table \ref{table:comparison}), our model's performance was compromised under most datasets (except MovieLens-IMDB). This indicated the semantic enrichment given by additional attributes for the column names, which were HXL-style tags, and this improvement helped our model achieve the best macro-F1 and AUC performance over other benchmarks.

 Compared to the public datasets, the HDXSM dataset contained significantly fewer values per schema. As a result, schema-based models like COMA, Cupid, and Similarity Flooding achieved higher F1/AUC scores than the distribution-based, value-embedded, and Jaccard-Levenshtein models, which rely more heavily on  column values.

\begin{figure}[h]
    \centering
    \begin{subfigure}[b]{0.38\textwidth}
        \includegraphics[width=\textwidth]{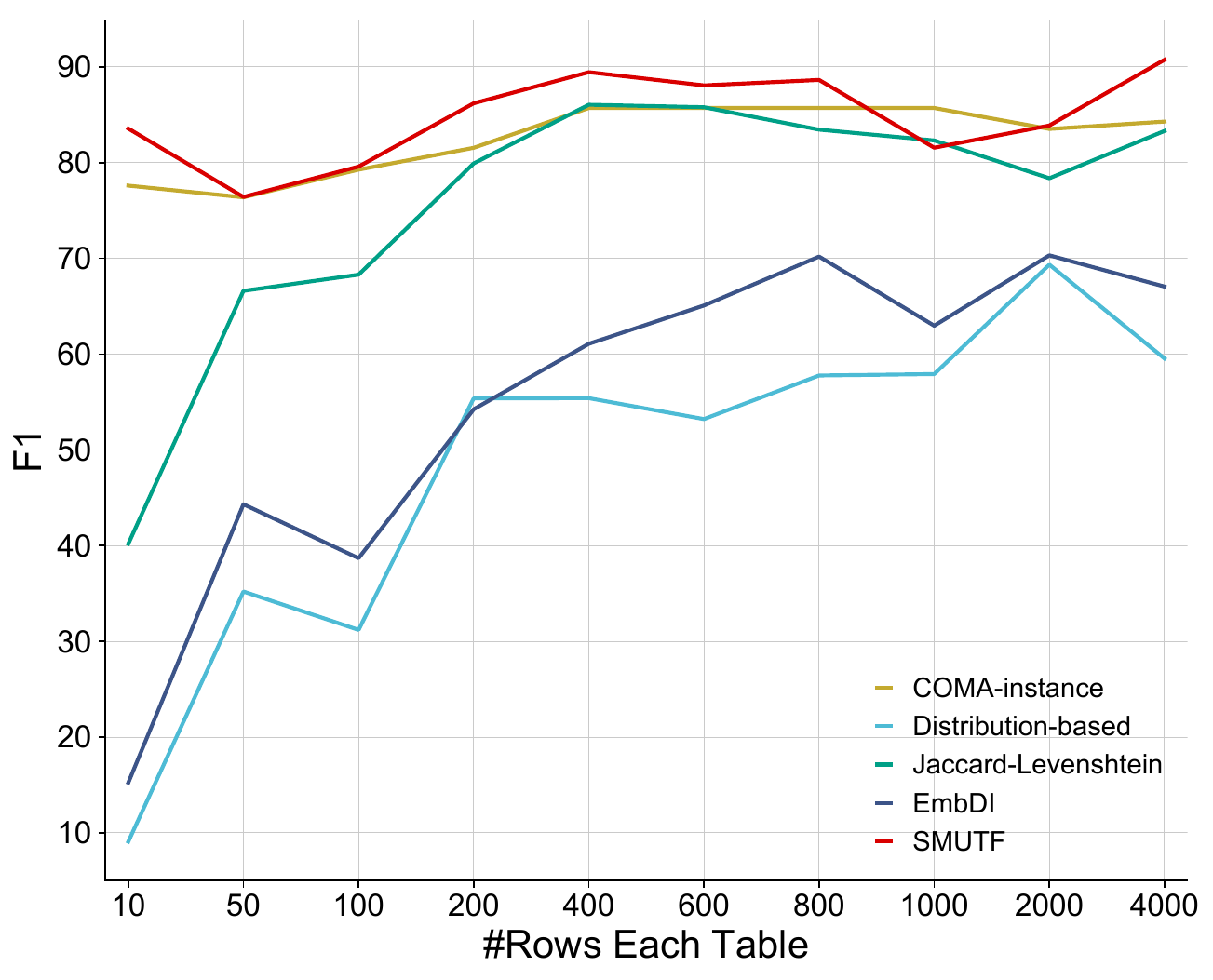}
        \caption{F1}
        \label{fig:pdf1}
    \end{subfigure}
    \hspace{1cm}
    \begin{subfigure}[b]{0.38\textwidth}
        \includegraphics[width=\textwidth]{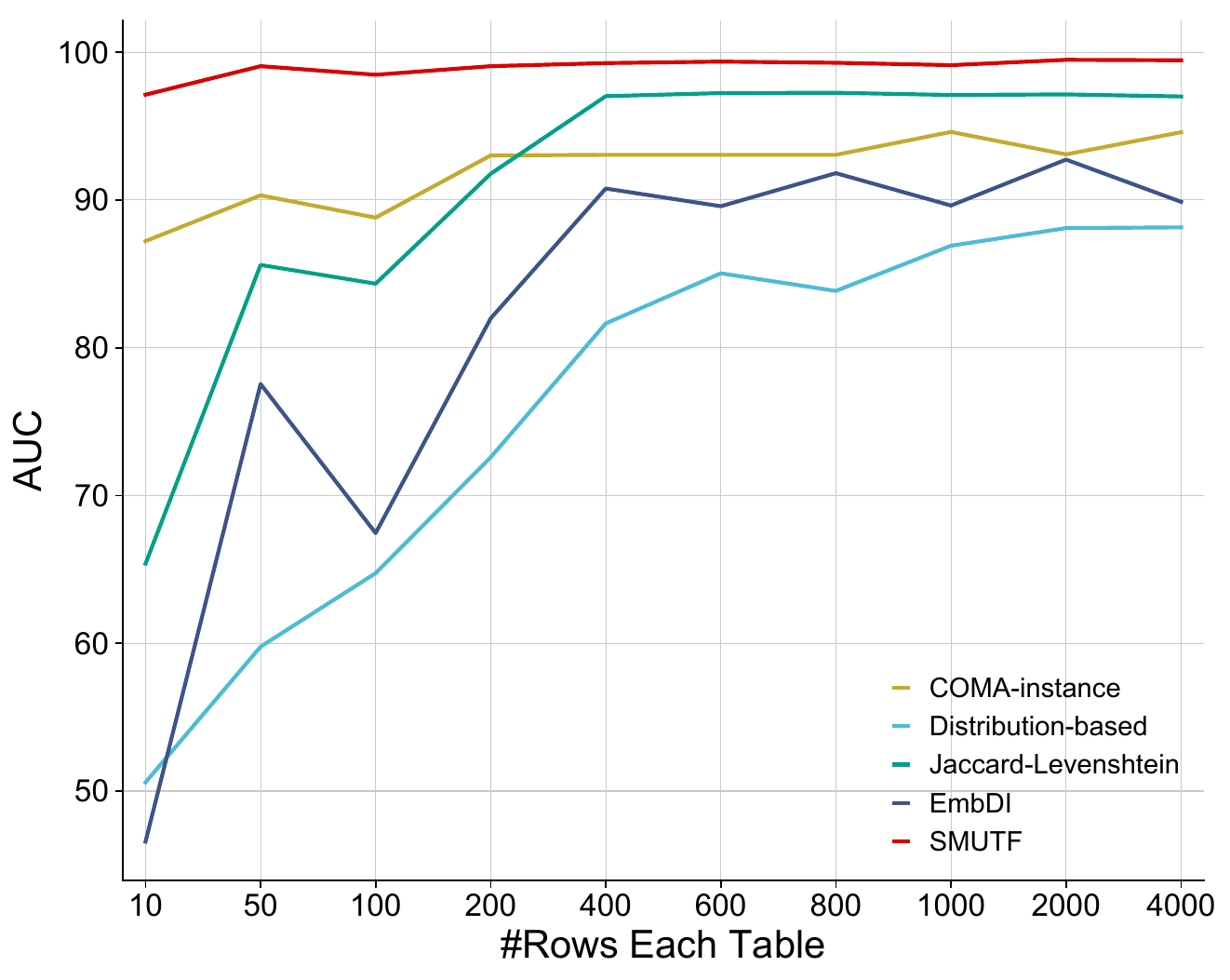}
        \caption{AUC}
        \label{fig:pdf2}
    \end{subfigure}
    \caption{Row Number Impact on Performance of Value-Based Methods}
    \label{fig:row_experiment}
    \vspace{-0.1in}
\end{figure}

In the case of the DI2KG dataset for Monitor and Camera categories, the performance of nearly all existing SM methodologies was suboptimal. Even the best-performing method, SMUTF, achieved an F1 score of merely 52.4 on the Camera dataset, and on Monitor, its F1 score was further reduced to 45.15. This underwhelming performance is attributable to a confluence of factors. On  the one hand, the Monitor and Camera datasets embody the most complex and in-depth technical attributes among all datasets examined. The datasets include numerous specialized concepts related to equipment, with corresponding data points that are extremely similar in nature and thus complicates discernment without prior domain knowledge. For instance, within the Camera category, there are three distinct types of resolution: "image resolution", "video resolution", and "sensor resolution" (often indicated in Megapixels). They all pertain to resolution and share similar data formats and values, represented either by dimensions or pixel count. Their close resemblance presents significant challenges for match prediction based on column names or values alone, highlighting the inherent limitations of such an approach. On the other hand, issues inherent to the DI2KG dataset itself, particularly concerning the mediated schema with instances of incorrect or incomplete matches, may contribute to the poor performance. Taking resolution as an illustrative example, an "image resolution" attribute from Website A might be matched to "megapixels" attribute on Website B, but when Website C presents an "image resolution" attribute or a similar one, it may not be flagged as a match. Previous studies have also identified the issue of duplicate attributes within the DI2KG dataset, necessitating regularized preprocessing to mitigate such complications \cite{obraczka2019knowledge}. It is possible that the challenges associated with the DI2KG dataset stem from its annotation process, as human annotators, when faced with a vast number of domain-specific tables, are unlikely to conduct a detailed inspection and judgement for each potential correspondence. Dataset creators may have resorted to automated methods for selection and preprocessing, yet the dataset currently lacks a detailed public disclosure of the annotation process and guidelines.  Unfortunately, although the Monitor and Camera datasets are significant challenges in the field of SM,  whether the reasons for poor SM performance stem from dataset quality issues or not is still unclear.

\begin{table}[h]
    \centering
    \setlength{\tabcolsep}{3.2mm}{
        \caption{The group of hyperparameters employed in the Tree-structured Parzen Estimator search.}
        \begin{tabular}{ll}
        \toprule
        Hyperparameter & Group \\
        \midrule
         learning\_rate & $[0.1,0.08,0.05,0.03]$ \\
         max\_depth & $[3,4,5,6,7,8,9,10]$ \\
         num\_round & $[100,200,300,400,500,600,700]$ \\
         \bottomrule
        \end{tabular}
        \label{tab:hyper}
    }
    \vspace{-0.2in}
\end{table}

\begin{table}[h]
    \centering
    \caption{Evaluating the performance of various sentence embedding models using the WikiData dataset.}
    \setlength{\tabcolsep}{2.8mm}{
    \begin{tabular}{lcccc}
    \toprule
    Model &  Precision & Recall & F1 & AUC \\
    \midrule
     Multilingual USE & 83.83 & 92.08 & 87.16 & 99.55 \\
     Multilingual MiniLM & 82.64 & 98.75 & 89.32 & \textbf{99.70} \\
     Multilingual MPNet & \textbf{84.72} & \textbf{100.00} & \textbf{91.39} & 99.53 \\
     \bottomrule
    \end{tabular}
    }
    \label{tab:sentence_embedding}
    \vspace{-0.1in}
\end{table} 

\begin{table}[h]
    \centering
    \caption{Evaluating the performance of various machine learning models using the WikiData dataset.}
    \setlength{\tabcolsep}{2.8mm}{
    \begin{tabular}{lcccc}
    \toprule
    Model &  Precision & Recall & F1 & AUC \\
    \midrule
     GaussianNB & 72.38 & 64.17 & 66.08 & 91.49 \\
     MLP & 84.38 & 93.13 & 87.71 & \textbf{99.54} \\
     LightGBM & 84.38 & 95.63 & 89.09 & 99.44 \\
     XGBoost & \textbf{84.72} & \textbf{100.00} & \textbf{91.39} & 99.53 \\
     \bottomrule
    \end{tabular}
    }
    \label{tab:ml_model}
    \vspace{-0.2in}
\end{table} 

\begin{table*}[htp]
\caption{Ablation studies on the components of SMUTF, which the dataset is WikiData and Camera. "Tag" means HXL-style Tag.}
\vspace{0.01in}
\setlength{\tabcolsep}{4mm}
\resizebox{\textwidth}{!}{%
\begin{tabular}{ccccccccrrll}
\hline
\multicolumn{2}{c}{Column Name Features} & \multicolumn{5}{c}{Value Features} & \multirow{2}{*}{HXL-style Tag} & \multicolumn{2}{l}{Wikidata} & \multicolumn{2}{l}{Camera} \\ \cline{1-7} \cline{9-12} 
Rule-based & Embedding & Data Type & Length & \multicolumn{1}{l}{Numerical} & \multicolumn{1}{l}{Character} & Embedding &  & \multicolumn{1}{c}{F1} & \multicolumn{1}{c}{AUC} & \multicolumn{1}{c}{F1} & \multicolumn{1}{c}{AUC} \\ \hline
 & \checkmark & \checkmark & \checkmark & \checkmark & \checkmark & \checkmark & \checkmark & \multicolumn{1}{l}{83.28} & \multicolumn{1}{l}{97.97} & 49.07 & 92.51 \\
\checkmark &  & \checkmark & \checkmark & \checkmark & \checkmark & \checkmark & \checkmark & 88.42 & 99.52 & 37.65 & 87.60 \\
\checkmark & \checkmark &  & \checkmark & \checkmark & \checkmark & \checkmark & \checkmark & 90.07 & 99.86 & 39.95 & 92.32 \\
\checkmark & \checkmark & \checkmark &  & \checkmark & \checkmark & \checkmark & \checkmark & 88.76 & 99.52 & 51.32 & 93.13 \\
\checkmark & \checkmark & \checkmark & \checkmark &  & \checkmark & \checkmark & \checkmark & 84.99 & 99.50 & 41.78 & 93.65 \\
\checkmark & \checkmark & \checkmark & \checkmark & \checkmark &  & \checkmark & \checkmark & 78.57 & 97.66 & 38.3 & 87.99 \\
\checkmark & \checkmark & \checkmark & \checkmark & \checkmark & \checkmark &  & \checkmark & 67.32 & 94.52 & 36.78 & 88.12 \\
\checkmark & \checkmark & \checkmark & \checkmark & \checkmark & \checkmark & \checkmark &  & 85.47 & 98.78 & 49.8 & 92.08 \\
\checkmark & \checkmark & \checkmark & \checkmark & \checkmark & \checkmark & \checkmark & \checkmark & \textbf{91.39} & \textbf{99.53} & \textbf{52.4} & \textbf{91.32} \\ \hline
\end{tabular}%
}
\label{tab:ablation}
\vspace{-0.1in}
\end{table*}

We also conducted an experiment on WikiData to assess the robustness of various value-based methods to changes in row count. For each sampled table from WikiData, we varied the row count from 10 to 4000 and reran our pattern-matching experiment. As shown in Figure \ref{fig:row_experiment}, the performance of the three methods, Jaccard-Levenshtein method, distribution-based approach, and EmbDI, significantly deteriorated when the row count was reduced to less than 100, indicating less robustness to row count variations. COMA-Instance also demonstrated satisfactory and stable results; however, in the majority of F1 scores and across all AUC metrics, SMUTF remained more robust than COMA-Instance. In experiments with over 1000 rows, SMUTF's performance in terms of F1 scores exhibited some fluctuations. This occurred because SMUTF, in its most vital part of column values' deep embedding, calculates only a random selection of 20 values per column. Consequently, a substantial increase in row count only affects other rule-based features. The observed fluctuations are due to the randomness inherent in the row sampling process within the experiments.

Compared to other benchmarks, similarity flooding has relatively high AUC scores over all datasets, and this indicates its good probabilistic capture of all matched pairs over non-matched pairs. The basic goal of similarity flooding is to compute the inter-node similarity between two graphs of database schema. Its core similarity propagation mechanism, where columns of two distinct graphs are similar when their adjacent columns are similar, helps the algorithm efficiently gain a global sense over all columns. Such a general focus may not result in a good F1 since the threshold value is hard to be chosen, but its AUC score that measures its general pair prediction capability is more impressive than others.

In a nutshell, our hybrid model that uses generative tags consistently outperformed traditional approaches in schema matching (SM) across various scenarios, surpassing schema-based methods by not relying solely on the linguistic similarity of names. Leveraging a pre-trained language model (PLM) encoder, our model automates textual similarity calculations without manually labeled tags. It outshines value-based methods by including schema information, making it more effective for schemas with fewer values in columns. The combination of schema and value features in our model strikes a balance, leading to superior performance in nearly all benchmark comparisons, as evidenced in Table \ref{table:comparison}.

\subsection{Hyperparamater Tuning}

We utilized the Tree Parzen Estimators (TPE) \cite{bergstra2011algorithms} method to optimize the hyperparameters of the XGBoost classifier from our predefined set (referenced in Table \ref{tab:hyper}). Additionally, we applied TPE to refine the hyperparameters of our baseline models, aligning with the configurations from Valentine research \cite{Koutras2020ValentineEM} to improve result quality. While the previously reported outcomes were obtained using a uniform configuration, fine-tuning the parameters for specific tasks has significantly improved the results.
\vspace{-0.1in}
\subsection{Ablation studies}\label{subsec:ablation}

In the ablation study, our goal is to examine how gradient boosting methods and sentence embedding technologies affect the performance of SM.

\subsubsection{The influence of the sentence embedding model}
We investigated the influence of various sentence embedding methods on our SM technique, comparing three pre-trained language models: Sentence Encoder \cite{yang2019multilingual}, MiniLM \cite{wang2020minilm}, MPNet \cite{song2020mpnet}. After applying knowledge distillation and training from the \cite{reimers-gurevych-2020-making} research team, we deployed the multilingual iterations of the models, with the findings detailed in Table~\ref{tab:sentence_embedding}. MPNet outshone other multilingual embeddings, likely because of its sophisticated architecture that merges features of both Masked and Permuted Language Models. Hence, MPNet was the chosen model for incorporation into SMUTF.

\subsubsection{The influence of the machine learning classification model}
The comparative evaluation of machine learning models on the WikiData dataset in Table~\ref{tab:ml_model} reveals XGBoost as the frontrunner, delivering the highest precision and recall, which implies the best F1 score. Notably, MLP with the highest AUC and LightGBM exhibit competitive precision. However, MLP falls short for the recall and F1 score. These results highlight XGBoost's exceptional ability to accurately predict relevant columns, positioning it as the optimal model for this dataset.

\renewcommand{\arraystretch}{1.5} %

\subsubsection{The influence of feature components}

Table \ref{tab:ablation} provides an analysis focusing on the effect of different components of SMUTF on SM performance, as evaluated on the WikiData dataset. In particular, we considered two main categories of features: Column Name Features and Value Features, both of which were evaluated using rule-based methods and embedding models. Additionally, we examined the influence of our novel HXL-style tagging system. 

In the results, our observations indicate that the ablation of any component within SMUTF precipitates a decline in its efficacy, signifying the contributory importance of each feature. The most significant impact arises from the removal of the Deep Embedding component within the Value Features category, with the F1 score plummeting to a mere 67.32. This substantial decrease underscores the criticality of value-based methods in SM and the insufficiency of relying solely on rule-based features for comprehensive value comparison. On the contrary, the component with the minimal influence is the Data Type Features, also within the Value Features category. The exclusion of this component results in a marginal F1 score reduction of 1.32, and intriguingly, the AUC exhibits a slight increase. This phenomenon can be primarily attributed to the relatively straightforward and superficial nature of data type judgments, which can be inferred through other rule-based and deep embedding features, as well as HXL-style tagging.

\section{Conclusion and Future Work}

We introduced SMUTF, a new method for SM in tabular data, which discerns dataset relationships through a composite strategy: creating HXL-style tags, rule-based feature extraction, deep embedding similarity, and XGBoost for similarity score prediction. This multi-faceted approach improves adaptability and schema alignment using a pre-trained model. Additionally, we presented the HDXSM Dataset, a substantial real-world SM dataset with 204 table pairs from the Humanitarian Data Exchange. Our evaluations against six benchmark methods showed that SMUTF has superior performance. Ablation studies confirmed the significant impact of each component on our method's effectiveness, particularly the value features. 

While SMUTF has demonstrated its strength as a system in SM tasks, we acknowledge that there are still many opportunities for improvement in the future work:

\begin{enumerate}
    \item \textit{Improving Generative Tagging:} Our novel introduction of generative tagging, inspired by the Humanitarian Exchange Language (HXL), proved beneficial to the SM process. We intend to further refine the tagging process by investigating more complex and dynamic tagging mechanisms that can better capture the semantics of columns in the data.
    \item \textit{Multi-modal SM:} SMUTF currently focuses on text-based tabular data. However, as we move towards increasingly complex data environments, multi-modal data such as images and videos are becoming more prevalent. Extending SMUTF to handle multi-modal data would increase its applicability.
    \item \textit{Leveraging Graph-based Models}: Our current methodology primarily relies on rule-based features, pretrained language models and machine learning model. However, schemas can naturally be represented as graphs, which allows for the use of recent advancements in graph neural networks for SM. Exploring graph-based models to improve SM could be a promising direction.

\end{enumerate}

\clearpage

\bibliographystyle{elsarticle-num}
\bibliography{custom}

\end{document}